\documentclass[conference,letterpaper]{IEEEtran}

\usepackage[utf8]{inputenc} 
\usepackage[T1]{fontenc}
\usepackage{url}
\usepackage{ifthen}
\usepackage{bm}
\usepackage{cite}
\usepackage{amssymb}
\usepackage[cmex10]{amsmath} 
\usepackage{cleveref}
\usepackage{bbm}
\usepackage{algorithm}
\usepackage{algorithmic}
\usepackage{graphicx}
\usepackage{booktabs}
\usepackage{caption}
\usepackage{subfig}
\usepackage[font=small]{caption}

\usepackage{xcolor}

\DeclareMathOperator*{\argmin}{arg\,min}
\DeclareMathOperator*{\argmax}{arg\,max}

\begin{document}
\newcommand{\ar}{\lambda} % arrival rate
\newcommand{\sr}{\mu} % service rate
\newcommand{\nums}{k} % number of servers
\newcommand{\fr}{\gamma} % fraction of total servers

\newcommand{\setc}{\mathcal{C}}
\newcommand{\tm}{\theta}
\newcommand{\ds}{N}
\newcommand{\sv}{\sigma^2}
\newcommand{\ibd}{\bm{I}_d}
\newcommand{\rd}{\mathbb{R}^d}
\newcommand{\bg}{\bm{g}}
\newcommand{\q}{q}
\newcommand{\hc}{q}
\newcommand{\ec}{M}
\newcommand{\scc}{m}
\newcommand{\ic}{k}
\newcommand{\bH}{\bm{H}}
\newcommand{\bF}{\bm{F}}
\newcommand{\icc}{{k'}}
\newcommand{\pof}{\textsc{Power-of-choice}}
\newcommand{\algoname}{\texttt{COIN}}

\newcommand{\prob}{\mathbb{P}}
\newcommand\tikzmark[1]{\tikz[remember picture,overlay] \coordinate (#1);}

\newcommand{\gf}{F}
\newcommand{\sgf}{\widetilde{\gf}}
\newcommand{\lff}{f}
\newcommand{\lf}{F_\ic}
\newcommand{\gl}{\mathcal{L}}
\newcommand{\hh}{\widehat{h}}

\newcommand{\sigd}{\sigma}
\newcommand{\sigl}{s}
\newcommand{\siglc}{\bm{s}}
\newcommand{\siglb}{\overline{\bm{s}}}
\newcommand{\cent}{\bm{c}}
\newcommand{\nc}{c}
\newcommand{\sw}{\alpha}
\newcommand{\dd}{\mathcal{D}}
\newcommand{\ddh}{\widehat{\mathcal{D}}}
\newcommand{\bA}{{\bm A}}
\newcommand{\bx}{{\bm x}}
\newcommand{\be}{{\bm e}}
\newcommand{\bb}{{\bm b}}
\newcommand{\bu}{{\bm u}}
\newcommand{\bdx}{\dot{\bm x}}
\newcommand{\bp}{{\bm \varpi}}
\newcommand{\bdp}{\dot{\bm p}}
\newcommand{\bq}{{\bm q}}
\newcommand{\bdq}{\dot{\bm q}}
\newcommand{\bX}{{\bm X}}
\newcommand{\bdX}{\dot{\bm X}}
\newcommand{\bQ}{{\bm Q}}
\newcommand{\bdQ}{\dot{\bm Q}}
\newcommand{\bP}{{\bm P}}
\newcommand{\bU}{{\bm U}}
\newcommand{\bV}{{\bm V}}
\newcommand{\bSigma}{{\bm \Sigma}}
\newcommand{\bdP}{\dot{\bm P}}
\newcommand{\by}{{\bm y}}
\newcommand{\bdy}{\dot{\bm y}}
\newcommand{\bv}{{\bm v}}
\newcommand{\bh}{{\bm h}}
\newcommand{\bI}{{\bm I}}
\newcommand{\bdv}{\dot{\bm v}}
\newcommand{\bw}{{\bm w}}
\newcommand{\bdw}{\dot{\bm w}}
\newcommand{\bt}{{\bm t}}
\newcommand{\bS}{{\bm S}}
\newcommand{\bsS}{{\bm \mathcal{S}}}
\newcommand{\bsD}{{\bm \mathcal{D}}}

\newcommand\inner[2]{\left\langle #1, #2 \right\rangle}
\newcommand\ind[1]{\mathbb{I}\left\{ #1 \right\}}
\newcommand\norm[1]{\left\| #1 \right\|_2}
\newcommand\ts[1]{\texttt{#1}}
\newcommand{\lmi}{\bW_i^*}
\newcommand{\tmi}{\bW_i^*}
\newcommand{\gm}{\widetilde{\bW}^*}
\newcommand{\kfi}{\widetilde{\bF}_i^{*}}
\newcommand{\tfi}{\bF_i(\bW_i^*)}
\newcommand{\vavg}{\bar{\bW}^*}
\newcommand{\gb}{\bm{g}}
\newcommand{\bmf}{\bm{f}}
\newcommand{\gc}{\Phi}
\newcommand{\wb}{\bm{w}}
\newcommand{\bbf}{\bm{f}}
\newcommand{\xb}{\bm{x}}
\newcommand{\wbh}{\widehat{\bm{w}}}
\newcommand{\wbhs}{\widehat{\bm{w}}^*}
\newcommand{\wbl}{\widetilde{\bm{w}}}
\newcommand{\zb}{\bm{z}}
\newcommand{\yb}{\bm{y}}
\newcommand{\xxb}{\bm{X}}
\newcommand{\hb}{\bm{H}}
\newcommand{\pb}{\bm{P}}
\newcommand{\mb}{\bm{M}}
\newcommand{\ib}{\bm{I}}
\newcommand{\eb}{\bm{e}}
\newcommand{\bd}{\mathcal{B}}
\newcommand{\bdat}{\mathcal{B}_\ic}
\newcommand{\diag}{\text{diag}}
\newcommand{\rr}{\mathbb{R}}
\newcommand{\ldat}{m_\ic}
\newcommand{\adat}{m}
\newcommand{\rdat}{p_k}
\newcommand{\rdati}{p_i}
\newcommand{\defeq}{\vcentcolon=}

\newcommand{\qw}{q_k}
\newcommand{\qsc}{\bm{q}}
\newcommand{\ip}{(i)}
\newcommand{\ti}{(t)}
\newcommand{\balpha}{\bm \alpha}
\newcommand{\tip}{(t+1)}
\newcommand{\tin}{(t-1)}
\newcommand{\midat}{\xi_{\ic}^{\ti}}
\newcommand{\midatc}{\xi_{\ic}^{\ip}}
\newcommand{\midatcc}{\xi_{\icc}^{\ip}}
\newcommand{\sgrad}{\bm{g}_\ic}
\newcommand{\hgrad}{h_\ic}
\newcommand{\grad}{G}
\newcommand{\wvar}{\sigma}
\newcommand{\lc}{\tau}
\newcommand{\lr}{\eta_t}
\newcommand{\lrt}{\eta_t^2}
\newcommand{\gri}{r}
\newcommand{\gr}{(r)}
\newcommand{\grp}{(r+1)}
\newcommand{\rcoi}{\hat{f}_{\btheta,i}}
\newcommand{\cm}{\btheta}
\newcommand{\substt}{\mathcal{A}}
\newcommand{\eval}{Z_k^{\ti}}
\newcommand{\owglb}{\overline{\wb}}
\newcommand{\hwglb}{\widehat{\wb}}
\newcommand{\wgopt}{\bm{w}^*}
\newcommand{\vglb}{\bm{v}}
\newcommand{\bW}{\bm{W}}
\newcommand{\btheta}{\bm{\theta}}
\newcommand{\gsto}{\bm{g}}
\newcommand{\ogsto}{\overline{\gsto}}
\newcommand{\hgsto}{\widehat{\gsto}}
\newcommand{\expt}{\mathbb{E}}
\newcommand{\vect}{\text{vec}}
\newcommand{\ba}{\bm{a}}
\newcommand{\bs}{\bm{s}}
\newcommand{\bG}{\bm{G}}
\newcommand{\bB}{\bm{B}}
\newcommand{\bz}{\bm{z}}
\newcommand{\iD}{\mathcal{D}}
\newcommand{\pktt}{s_{\ic}^{\ti}}
\newcommand{\zktt}{Z_{\ic}^{\ti}}
\newcommand{\qktt}{q_k}
\newcommand{\E}[1]{\mathbb{E}\left[{#1}\right]}
\newcommand{\Eg}[2]{\mathbb{E}_{#1}\left[{#2}\right]}
\newcommand{\tft}{\widetilde{\bm{f}}}
\newcommand{\nott}{\Sigma_{\ti}}
\newcommand{\dkr}{\frac{\ldat}{\adat}}
\newcommand{\dkrl}{l_{\ic}}
\newcommand{\nbr}[1]{{\left\| {#1} \right\|}}
\newcommand{\pir}{\pi_{\text{rand}}}
\newcommand{\pire}{\pi_{\text{rand-r}}}
\newcommand{\pia}{\pi_{\text{afl}}}
\newcommand{\pipd}{\pi_{\text{pow-d}}}
\newcommand{\pipr}{\pi_{\text{pow-r}}}
\newcommand{\piprc}{\pi_{\text{cpow-d}}}
\newcommand{\piprr}{\pi_{\text{rpow-d}}}
\newcommand{\rhb}{\overline{\rho}}
\newcommand{\rht}{\widetilde{\rho}}
\newcommand{\tgam}{\Gamma}

% \DeclareMathOperator*{\mint}{min} % thin space, limits on side in displays
% \DeclareMathOperator*{\maxt}{max} % thin space, limits on side in displays
 % thin space, limits on side in displays
% \DeclareMathOperator*{\argmax}{arg\,max} % thin space, limits on side in displays
% \DeclareMathOperator*{\argsort}{arg\,sort}
% \DeclarePairedDelimiter\ceil{\lceil}{\rceil}
% \DeclarePairedDelimiter\floor{\lfloor}{\rfloor}

\newcommand{\st}{\mathcal{S}^{\ti}}
\newcommand{\stt}{\mathcal{S}^{(t+1)'}}
\newcommand{\exptset}{\mathbb{E}_{\st}}
\newcommand*{\Comb}[2]{{}^{#1}C_{#2}}
\renewcommand{\mod}{~\mathrm{mod}~}
\title{Erasure Coded Neural Network Inference \\ via Fisher Averaging}

\author{%
  \IEEEauthorblockN{Divyansh Jhunjhunwala$^*$, Neharika Jali$^*$, and Gauri Joshi}
  \IEEEauthorblockA{Department of Electrical and Computer Engineering \\
Carnegie Mellon University\\
Pittsburgh, PA, USA\\
Email: \{djhunjhu, njali, gaurij\}@andrew.cmu.edu}
  \and
  \IEEEauthorblockN{Shiqiang Wang}
  \IEEEauthorblockA{IBM Research\\ 
Yorktown Heights, NY, USA\\
Email: wangshiq@us.ibm.com}
}

\maketitle

\def\thefootnote{*}\footnotetext{Denotes equal contribution}

\begin{abstract}
Erasure-coded computing has been successfully used in cloud systems to reduce tail latency caused by factors such as straggling servers and heterogeneous traffic variations. A majority of cloud computing traffic now consists of inference on neural networks on shared resources where the response time of inference queries is also adversely affected by the same factors. However, current erasure coding techniques are largely focused on linear computations such as matrix-vector and matrix-matrix multiplications and hence do not work for the highly non-linear neural network functions. In this paper, we seek to design a method to code over neural networks, that is, given two or more neural network models, how to construct a coded model whose output is a linear combination of the outputs of the given neural networks. We formulate the problem as a KL barycenter problem and propose a practical algorithm \ts{\algoname} that leverages the diagonal Fisher information to create a coded model that approximately outputs the desired linear combination of outputs. We conduct experiments to perform erasure coding over neural networks trained on real-world vision datasets and show that the accuracy of the decoded outputs using \ts{\algoname} is significantly higher than other baselines while being extremely compute-efficient. 
\end{abstract}

\section{Introduction} \label{sec:introduction}

Modern machine learning (ML) jobs are deployed on large-scale cloud-based computing infrastructure. With training being a one-time event, an overwhelming majority of cloud computing traffic now constitutes ML inference jobs, in particular, inference on neural network models. Inference queries are highly time-sensitive because delays and time-outs can directly impact the quality of service to users. However, the ML models in question are often foundation models trained on diverse large-scale datasets and fine-tuned for different specific downstream tasks \cite{bommasani2021foundationmodels}. Due to the size and computational complexity of these models, there can be significant variations in the time taken to process inference queries. Guaranteeing low inference latency is all the more challenging because applications now host multiple neural network models on the same shared infrastructure. Issues such as resource contention in multi-tenant clusters \cite{xu2013Bobtail}, network constraints \cite{crankshaw2017Clipper} or hardware unreliability \cite{ananthanarayanan2010Reining} can result in straggling servers. The straggler problem is even worse in ensemble inference scenarios \cite{ sagi2018ensemble}, where the desired inference output is a combination of the outputs of an ensemble of models, because delays in the output of any one model in the ensemble can bottleneck the entire query. Heterogeneous traffic for different models hosted on the same infrastructure can also be a major issue affecting latency. In many applications, inference queries are routed to one of the several expert models based on the features of the query \cite{jordan1994hierarchical,fedus2022review}. For example, an object detection application may use specialized models for indoor and outdoor images or day and nighttime images. In such scenarios, the query traffic for each model can vary unpredictably over time \cite{googleTrends} and it can be negatively correlated across models, i.e., if the volume of queries to one model is high, it is low for another model. 

Techniques to handle stragglers and unpredictable traffic variations include launching redundant queries \cite{dean2008mapreduce, zaharia2010spark}, also referred to as speculative execution and replication of models to meet the highest possible traffic demand. These lead to inefficient execution of queries and idling of resources respectively. Erasure coding, which is a generalization of replication, is an effective solution for straggler mitigation in matrix computations \cite{ananthanarayanan2013effective, joshi2018synergy, mallick2020rateless} and handling heterogeneous traffic \cite{aktas2017service, anderson2018service, choudhury2022erasurecode}. As an illustrative example, we look at a linear inference task of matrix computation consisting of models $\mathbf{A}_1, \mathbf{A}_2, \dots, \mathbf{A}_N$. Consider an ensemble inference query, represented by vector $\bx$, that requires outputs of all the $N$ models and can be bottlenecked by straggling of any one of the outputs. If we create coded linear combinations of the $N$ models, they can allow us to recover the $N$ outputs even when one of the uncoded models slows down. To see the benefit of coding in handling heterogeneous traffic, consider an application scenario that routes the query $\bx$ to one of the models $\mathbf{A}_1, \mathbf{A}_2, \dots \mathbf{A}_N$, and the output required to be computed is $\mathbf{A}_i \bx$. If the server storing $\mathbf{A}_i$ is congested or fails, the query routed to two servers one with $\mathbf{A}_j$ and one with coded $\mathbf{A}_i + \mathbf{A}_j$ and the outputs can be combined to recover the desired output $\mathbf{A}_i \bx$. The coded server thus acts as a flexible model that can effectively be used to serve queries of both types or enable retrieval if one of the servers is slow. 

\textbf{Main Contributions.}

A key missing element in previous works is that erasure coding is inherently linear and does not work for non-linear functions. Thus, it cannot be directly used for neural network inference. In this work, we consider the question of how to erasure code neural networks. In \Cref{sec:problemFormulation}, we define the coding objective that seeks to construct a coded network whose output is a linear combination of two or more neural networks. In \Cref{sec:algorithm} we reduce our coding objective to the equivalent $\mathrm{KL}$ barycenter problem and propose a practical solution \ts{\algoname} that approximately produces the desired linear combination of outputs. Finally in \Cref{sec:experiments}, we measure the accuracy of the decoded outputs using \ts{\algoname} on neural networks trained on real-world vision datasets and highlight its improvement over several competing baselines. To the best of our knowledge, \ts{\algoname} is one of the first works to do erasure coding for non-linear neural network functions in a compute-efficient manner.

\section{Related Work} \label{sec:relatedWork}
Erasure codes have been extensively studied and applied to distributed storage and computing. Effective solutions for both straggler mitigation and latency reduction use principles of replication and erasure codes \cite{ananthanarayanan2013effective, joshi2018synergy, mallick2020rateless}. Other works including \cite{karakus2017encoded, tandon2017gradient, halbawi2018gdrscodes, dutta2018polydot} use erasure codes for straggler-resistant computation for convex optimization and gradient descent. In \cite{aktas2017service, anderson2018service, choudhury2022erasurecode}, the authors proposed the idea of using coded servers to handle variations in skewed traffic for efficient reduction of the response time of queries. However, the codes in the above works are for linear function computations and do not work in general for non-linear functions like neural networks.  

There is very little work on employing erasure codes for non-linear functions and neural networks, some of which include the following relevant work. \cite{mallick2022ratelessnonlinear} decomposes non-linear functions into inexpensive linear functions and proposes rateless sum-recovery codes to alleviate the problem of stragglers in distributed non-linear computations. For inference on an ensemble of neural networks, \cite{kosaian2019parity} proposes learning a `parity' network that is trained to transform erasure-coded queries into a form that enables a decoder to reconstruct slow or failed predictions. A major drawback of this, however, is that it requires training the parity model from scratch, which is expensive in both compute and data requirements. Another orthogonal line of work aims to learn the encoder and decoder neural networks that enable erasure coding in communication over noisy channels \cite{kim2018communication, kim2018deepcode}. The goal of this line of work is different and complementary to ours -- they construct codes using deep neural networks while we are using codes to improve the reliability and latency performance of neural network inference.

A recent relevant line of work in ML has investigated the problem of `model fusion', i.e., combining the weights of two or more independent neural networks into a single network that broadly speaking inherits the properties of the fused networks \cite{wortsman2022model, IlharcoRWSHF23, Jin0P023, matena2022merging}. In this aspect, model merging is closer to multi-task learning \cite{caruana1997multitask} where the goal is to learn a single model that can perform well on all tasks. Our goal on the other hand is strictly to produce a model whose output is a linear combination of the given neural networks; we do not care about the performance of the coded model on individual tasks. The closest work to ours is \cite{matena2022merging} which also proposes to use the diagonal Fisher when merging models. Nonetheless, we believe our motivation for using the Fisher information for erasure coding is novel as discussed in \Cref{sec:problemFormulation} along with our experiments in \Cref{sec:experiments} which show that our proposed approach significantly improves decoding accuracy compared to approaches which are adopted from model merging literature. We discuss other related works on model merging in \Cref{sec:experiments}.

\section{Problem Formulation and Preliminaries} \label{sec:problemFormulation}
In the rest of the paper, we use lowercase bold letters, e.g. $\bx$, to denote vectors and use $x_i$ to denote the $i$-th element of the vector $\bx$. We use $\norm{\cdot}$ to denote the $L_2$ norm, $\mathbb{R}$ to denote the set of real numbers and $[N]$ to denote the set of numbers $\{1,2,\dots,N\}$. 
Probability density functions are represented by $p(\bx)$. We use $\mathrm{KL}(p(\bx)||q(\bx))$ to denote the KL divergence between two densities $p(\bx)$ and $\q(\bx)$.

\textbf{Multi-Model Inference Setup.} We consider a multi-model inference setup with $N$ neural networks denoted by $f_{\btheta_1}(\bx), f_{\btheta_2}(\bx),\dots,f_{\btheta_N}(\bx)$ where $\btheta_i \in \mathbb{R}^d$ parameterizes the weights of the $i$-th neural network. Each neural network takes an input $\bx \in \mathbb{R}^s$ (e.g., an image) and produces an output $\by \in \mathbb{R}^K$ (e.g., image label). To simplify our discussion and also allow comparison with other baseline methods in \Cref{sec:experiments}, we assume that these neural networks have the same architecture. 

\textbf{Erasure Coding Objective.}

We consider a class of erasure codes called systematic maximum distance separable (MDS) codes \cite{berkelamp1968algebraic,cover2006elements} that take $N$ source symbols ($N$ neural networks in our case) and produce $N_c$ coded symbols such that using the $N_c$ coded symbols and any subset of $(N-N_c)$ source symbols, we can recover the other $N_c$ source symbols. 
In this paper, we focus on the $N_c = 1$ case and leave extensions to general $k$ as future work. That is, given $N$ neural networks $f_{\btheta_1}(\bx),f_{\btheta_2}(\bx),\dots,f_{\btheta_N}(\bx)$, our goal is to produce a coded neural network $f_{\cm}(\bx)$ that can be used to recover the output of $f_{\btheta_i}(\bx)$ for any $i$ using the output of the remaining $(N-1)$ neural networks. To do so, we want to express $f_{\cm}(\bx)$ as a convex combination of $f_{\btheta_i}(\bx)$'s:
\begin{align}
\textstyle
    f_{\cm}(\bx) \approx \sum_{i=1}^N\beta_i f_{\btheta_i}(\bx)
\end{align}
where $\beta_i > 0, \sum_{i=1}^N \beta_i = 1$ are the coding weights.

Given such a coded neural network $f_{\cm}(\bx)$, it is easy to approximately recover $f_{\btheta_i}(\bx)$ using the other $N-1$ networks $\{f_{\btheta_j}(\bx)\}_{j=1,j\neq i}^N$ as follows,

\begin{align}
\label{eq:decoded_output}
    \rcoi(\bx) = \frac{1}{\beta_i}\left(f_{\cm}(\bx) - \sum_{j=1, j \neq i}^N \beta_j f_{\btheta_j}(\bx) \right) \approx f_{\btheta_i}(\bx).
\end{align}

The quality of the decoded approximation can be measured by the mismatch between $f_{\btheta_i}(\bx)$ and $\rcoi(\bx)$ for all $\bx \in \mathbb{R}^s$ and $i \in [N]$, which we define using the squared loss function: 
\begin{align}
\label{eq:obj_func}
    L(\cm) 
    &= \frac{1}{2N}\Eg{\bx}{\sum_{i=1}^N {\norm{f_{\btheta_i}(\bx) - \rcoi(\bx)}^2}}\\
    & = \frac{\bar{\beta}}{2}\Eg{\bx}{\norm{f_{\btheta}(\bx) - \sum_{i=1}^N \beta_i f_{\cm_i}(\bx) \label{eq:obj_simplified}
}^2},
\end{align}
where the last equality follows from substituting \Cref{eq:decoded_output} in \Cref{eq:obj_func} and defining $\bar{\beta} = (\sum_{i=1}^N 1/\beta_i^2)/N$. 
Since the distribution $q(\bx)$ over $\bx$ is typically unknown, we only assume access to $P$ samples $\bx_1,\bx_2,\dots,\bx_P$ drawn from $q(\bx)$. 
 
Given these $P$ samples and the neural networks $f_{\btheta_1}(\bx), f_{\btheta_2}(\bx),\cdots, f_{\btheta_n}(\bx)$, we can define the following empirical coding loss
\begin{align}
     \hat{L}(\cm)
     &= \frac{\bar{\beta}}{2P} \sum_{l=1}^P \norm{f_{\btheta}(\bx_l) - \sum_{i=1}^N \beta_i f_{\cm_i}(\bx_l) 
}^2. \label{eq:empirical_simplified}
\end{align}
We now discuss a baseline approach to minimize the empirical objective, followed by our proposed approach in \Cref{sec:algorithm}.

\textbf{Ensemble Distillation Baseline.} 

From \Cref{eq:empirical_simplified}, we see that we want the output of our coded neural network $f_{\btheta}(\bx)$ to match the output of the `ensemble' of neural networks given by $\sum_{i=1}^N \beta_i f_{\btheta_i}(\bx_l)$. This idea has been well studied in the context of ensemble distillation \cite{hinton2015distillknow, lin2020ensemble,freitag2017ensemble} where the goal is to distill the knowledge from an ensemble of models or `teachers' into a single model or `student'.

Treating the output $\sum_{i=1}^N \beta_i f_{\btheta_i}(\bx)$ as a pseudo-label $\hat{\by}_l$ for every $l \in [P]$, we see that our objective becomes exactly the same as squared loss regression and can be optimized with standard gradient-based techniques. However, there are some drawbacks to this approach. Firstly performing such a gradient based optimization step imposes a significant computation cost. Secondly, it is not easy to modify the coded network to account for changes in the coding weights $\beta_i$'s or add a new neural network $f_{\btheta_{N+1}}(\bx)$ to our coding setup. We would need to re-train the coded network in such cases. Lastly, in the case where the number of samples $P$ is small, there is a serious risk of overfitting. We demonstrate this in our experiments where we show that the coded network obtained via ensemble distillation generalizes poorly for samples outside of the training set. Standard regularization techniques such as early stopping and weight decay are also unable to help with the overfitting as we show in the Appendix.

%%%%%%%%%%%%%%%%%%%%%%%%%%%%
%%%%%%%%%%%%%%%%%%%%%%%%%%%%

\section{\ts{\algoname}:
COded Inference of Neural networks} \label{sec:algorithm}
In this section we show how the problem of minimizing the objective in \Cref{eq:obj_func} can be reformulated to get an equivalent problem known as the KL barycenter problem \cite{ClaiciYG020}. Next we discuss how to get an approximate solution to the KL barycenter problem in our setup and how we can practically implement this solution.

\textbf{Neural Network as a Statistical Model.}  To motivate our proposed solution, we use the idea of a neural network as a parameterized statistical model that defines a probability density function $p_{\btheta}(\bx,\by)$ over all input-label pairs $(\bx,\by)$ in $\mathbb{R}^{s \times K}$. In particular, we define $p_{\btheta}(\bx,\by) = q(\bx)p_{\btheta}(\by|\bx) = q(\bx)\exp\left(-\norm{\by-f_{\btheta}(\bx)}^2\right)/\sqrt{2\pi}$ where $\bx \sim q(\bx)$ is the input distribution over $\mathbb{R}^s$, which is independent of parameters $\btheta$. This is a standard idea in statistical learning that draws an equivalence between minimizing the squared loss and maximizing the log likelihood of the observed data under a Gaussian model since $-\log p_{\btheta} (\bx,\by) = \norm{f_{\btheta}(\bx) - \by}^2 + c$, where $c$ is some constant which does not depend on $\btheta$. 

\textbf{Reduction to KL Barycenter Problem.}  Expanding the norm in \Cref{eq:obj_simplified} and since $\sum_{i=1}^N \beta_i = 1$, we get
\begin{align}
L(\cm)
& = \frac{\bar{\beta}}{2} \underbrace{\sum_{i=1}^N \bar{\beta}_i \Eg{\bx}{\norm{f_{\btheta_i}(\bx) - f_{\cm}(\bx)}^2}}_{L_1(\cm)} \nonumber \\
& - \frac{\bar{\beta}}{2}\sum_{i=1}^N\sum_{j=1,j \neq i}^N \beta_i \beta_j\Eg{\bx}{\norm{f_{\btheta_i}(\bx) - f_{\btheta_j}(\bx)}^2}\label{eq:simplified_obj_1}
\end{align}
where $\bar{\beta_i} = \beta_i \sum_{j=1}^N  \beta_j$. Since the second term in \Cref{eq:simplified_obj_1} does not depend on the coded model's parameters $\cm$, we focus on minimizing just the first term $L_1(\cm)$: 
\begin{align}
   L_1(\cm) &= \frac{1}{2}\sum_{i=1}^N \bar{\beta}_i \Eg{\bx}{\norm{f_{\btheta_i}(\bx) - f_{\cm}(\bx)}^2}\\
   & = \sum_{i=1}^N \bar{\beta}_i\Eg{\bx}{\mathrm{KL}(p_{\btheta_i}(\by|\bx)||p_{\cm}(\by|\bx))} \label{eq:kl_simplify}\\
   & = \sum_{i=1}^N \bar{\beta}_i \mathrm{KL}(p_{\btheta_i}(\bx, \by)||p_{\cm}(\bx, \by)) \label{eq:kl_more_simplify}
\end{align}
where \Cref{eq:kl_simplify} follows from our definition of $p_{\btheta}(\by|\bx)$ above and uses the fact that $\mathrm{KL}(\mathcal{N}(\bm{\mu}_1,\bm{\Sigma})|| \mathcal{N}(\bm{\mu}_2,\bm{\Sigma})) = \norm{\bm{\mu}_1 - \bm{\mu}_2}^2/2$. Thus we see that minimizing $L_1(\cm)$ is equivalent to finding the density function $p_{\cm} (\bx,\by)$ that is a \textit{weighted average} (in the KL divergence sense) of the density functions $p_{\btheta_i}(\bx,\by)$ with weights proportional to $\bar{\beta}_i$. This is known as the KL barycenter problem and has been studied in previous work in the context of clustering \cite{banerjee2005clustering} and model-fusion \cite{ClaiciYG020}.

\textbf{Solving the KL Barycenter Problem.}
In the case where $p_{\btheta_i}(\bx,\by)$ belongs to the exponential family of distributions with natural parameter $\btheta$, it is known that there exists an analytical expression for the parameters $\cm$ of the distribution $p_{\cm}(\bx,\by)$ that minimizes \Cref{eq:kl_more_simplify} \cite{ClaiciYG020}. However, this is not the case in our setup because $p_{\btheta}(\by|\bx)$ in \Cref{eq:kl_simplify} is Gaussian with respect to $f_{\btheta}(\cdot)$ and not $\btheta$ itself. Thus, we need to resort to some approximations to get a analytical solution. We use the following approximation for the KL divergence between $p_{\btheta_i}(\bx,\by)$ and $p_{\btheta}(\bx,\by)$ \cite{martens2020new},
\begin{align} \label{eq:klApprox}
    \mathrm{KL}(p_{\btheta_i}(\bx,\by)||p_{\cm}(\bx,\by)) \approx (\btheta-\btheta_i)^{\top}F_{\btheta_i}(\btheta-\btheta_i)
\end{align}
where $F_{\btheta_i}$ is the Fisher information matrix of $\btheta_i$ defined as follows
\begin{align}
    F_{\btheta_i} &= \Eg{\bx}{\Eg{\by \sim p_{\btheta}(\cdot|\bx)}{\log p_{\btheta}(\bx,\by) \nabla_{\btheta} \log p_{\btheta}(\bx,\by)^{\top}}}_{\btheta = \btheta_i}\\
    & = \Eg{\bx}{\nabla_{\btheta}f_{\btheta}(\bx)\nabla_{\btheta}f_{\btheta}(\bx)^{\top}}_{\btheta = \btheta_i}
\end{align}
This approximation comes from treating $\mathrm{KL}(p_{\btheta_i}(\bx,\by)||p_{\btheta}(\bx,\by))$ as a function of $\btheta$ and taking a second order Taylor expansion around $\btheta_i$ (zeroth and first order terms are zero). As is the case with Taylor expansions, the quality of the approximation degrades as the distance $\norm{\btheta - \btheta_i}^2$ increases. To capture this we also propose to add a penalty term $\lambda\norm{\btheta - \btheta_i}^2, \lambda > 0$ to this approximation. With this, our new objective $G(\btheta)$ is given by,
\begin{align}
    L_{1}(\btheta) &\approx G(\btheta) \\
    &= \sum_{i=1}^N \bar{\beta}_i(\btheta - \btheta_i)^{\top}F_{\btheta_i}(\btheta - \btheta_i) + \lambda\sum_{i=1}^N \bar{\beta}_i\norm{\btheta - \btheta_i}^2
\end{align}
We see that $G(\btheta)$ is a strongly convex function whose global minimizer is given by,
\begin{align}
\label{eq:global_minimizer}
    \btheta^* & = \argmin_{\btheta \in \mathbb{R}^d} G(\btheta) \nonumber \\
    & = \left[ \sum_{i=1}^N\bar{\beta}_i(F_{\btheta_i} + \lambda I)\right]^{-1}\sum_{i=1}^N\bar{\beta}_i(F_{\btheta_i}+ \lambda I)\btheta_i.
\end{align}
Thus given the parameters of our uncoded networks $\btheta_i$, \Cref{eq:global_minimizer} outlines how in theory, we can compute $\btheta^*$ such that $f_{\btheta^*}(\bx) \approx \sum_{i=1}^N \beta_i f_{\btheta_i}(\bx)$. 

\begin{figure}
    \centering
    \includegraphics[width=0.4\textwidth]{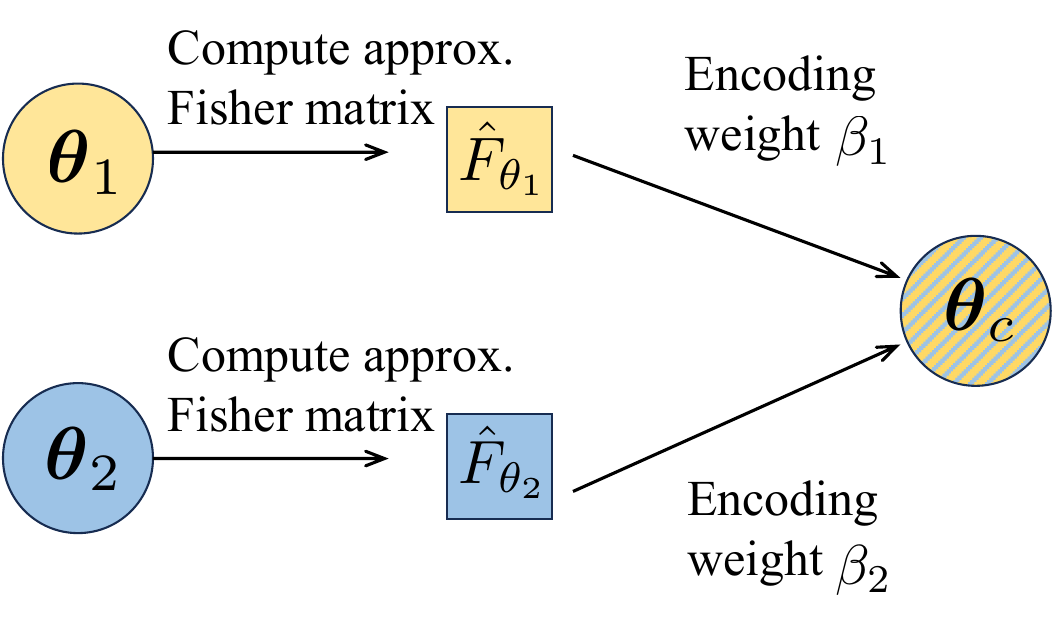}
    \caption{Illustration of how our proposed method \algoname~ (see Algorithm~\ref{algo1}) computes the coded model's parameters $\btheta_c$ such that its output $f_{\theta_c}(\bx) \approx \beta_1 f_{\theta_1}(\bx) + \beta_2 f_{\theta_2}(\bx)$, a linear combination of the outputs of $f_{\theta_1}(\bx)$ and $f_{\theta_2}(\bx)$. Unlike ensemble distillation, the parameters $\btheta_c$ are computed without requiring training the model from scratch.}
    \label{fig:codingDiagram}
\end{figure}

\setlength{\textfloatsep}{1em}%
\begin{algorithm} [t]
\caption{\ts{\algoname} }\label{algo1}

\begin{algorithmic}[1]
\STATE {\bfseries Input:} neural networks $f_{\btheta_1}(\bx),f_{\btheta_2}(\bx),\dots,f_{\btheta_N}(\bx)$ to be coded, coding weights $\beta_1,\beta_2,\dots,\beta_N$, input data samples $\bx_1,\bx_2,\dots,\bx_P$, penalty parameter $\lambda$

\STATE {\bfseries For $i \in [N]$ do:}
\STATE \hspace*{1em} Compute $\bar{\beta}_i = \beta_i \sum_{j=1}^N \beta_j$
\STATE \hspace*{1em} Compute \\ \quad \quad $\hat{F}_{\btheta_i} = \mathrm{diag}\left(\frac{1}{P}\sum_{l=1}^P [\nabla_{\btheta} f_{\btheta}(\bx_l)\nabla_{\btheta} f_{\btheta}(\bx_l)^{\top}]_{\btheta = \btheta_i}\right)$
\STATE Return  $\btheta_c = ( \sum_{i=1}^N \bar{\beta}_i(\hat{F}_{\btheta_i} + \lambda I))^{-1}\sum_{i=1}^N\bar{\beta}_i(\hat{F}_{\btheta_i}+ \lambda I)\btheta_i.$

\end{algorithmic} 
\end{algorithm}

\textbf{Practical Solver.} In practice, computing the exact Fisher $F_{\btheta_i}$ for all $i \in [N]$ in \Cref{eq:global_minimizer} is challenging since it involves $\mathcal{O}(d^2)$ operations, with $d$ being in the order of millions for neural networks. Also recall that we only have access to $P$ samples $\bx_1,\bx_2,\dots,\bx_P$ from the distribution $q(\bx)$. Thus in order to get a fast and tractable solution, we propose to approximate the true Fisher $F_{\btheta_i}$ with the diagonal of the empirical Fisher as done in several other works dealing with computing the Fisher \cite{kirkpatrick2017overcoming,lecun1989optimal}. In other words we have,
\begin{align}
    F_{\btheta_i} \approx \hat{F}_{\btheta_i} = \mathrm{diag}\left(\frac{1}{P}\sum_{l=1}^P [\nabla_{\btheta} f_{\btheta}(\bx_l)\nabla_{\btheta} f_{\btheta}(\bx_l)^{\top}]_{\btheta = \btheta_i}\right).
\end{align}
With this approximation, the parameters of our coded model $\btheta_c$ are given by,
\begin{align}
      \btheta_c = ( \sum_{i=1}^N \bar{\beta}_i(\hat{F}_{\btheta_i} + \lambda I))^{-1}\sum_{i=1}^N\bar{\beta}_i(\hat{F}_{\btheta_i}+ \lambda I)\btheta_i.
\end{align}
We find that using the diagonal Fisher is sufficient to provide a consistent improvement over other coding baselines including ensemble distillation as shown in \Cref{sec:experiments}. Furthermore it also overcomes other limitations of the ensemble distillation baseline - it is simple and cheap to compute, effectively taking only $O(d)$ operations and can be modified easily to incorporate changes in the coding weights $\beta_i$ or $\btheta_i$ without needing an expensive retraining step. It has also been shown that the empirical approximation of the Fisher is sample-efficient \cite{singh2020woodfisher}; we only choose $P$ to be about $200$ to get a good approximation to the true Fisher in our experiments. 

\begin{table*}[h]
\centering
\caption{Normalized Decoding Accuracy results when coding over experts trained on different partitions of the same dataset. \ts{\algoname} shows a significant improvement in performance compared to baselines while being compute-efficient.}
\color{black}

\begin{tabular}{lccc|ccc|ccc}
\toprule
&
\multicolumn{3}{c}{MNIST} &
\multicolumn{3}{c}{FashionMNIST} &
\multicolumn{3}{c}{CIFAR10}\\
\cmidrule{2-10}
Algorithm & 
\multicolumn{1}{c}{Split 1} &
\multicolumn{1}{c}{Split 2} &
\multicolumn{1}{c}{Avg.} &
\multicolumn{1}{c}{Split 1} &
\multicolumn{1}{c}{Split 2} &
\multicolumn{1}{c}{Avg.} &
\multicolumn{1}{c}{Split 1} &
\multicolumn{1}{c}{Split 2} &
\multicolumn{1}{c}{Avg.}\\
\midrule
Vanilla Averaging \cite{wortsman2022model}
& $95.61$
& $83.52$
& $89.56$
& $98.75$
& $92.21$
& $95.48$
& $94.28$
& $86.09$
& $90.19$\\
Task Arithmetic \cite{IlharcoRWSHF23}
& $96.82$
& $83.56$
& $90.19$
& $98.73$
& $92.21$
& $95.47$
& $95.19$
& $86.70$
& $90.94$\\
RegMean \cite{Jin0P023}
& $95.36$
& $83.91$
& $89.63$
& $96.50$
& $89.37$
& $92.93$
& $91.05$
& $85.05$
& $88.05$\\
Ensemble Distillation
& $97.19$
& $97.06$
& $97.12$
& $92.98$
& $84.63$
& $88.80$
& $85.33$
& $90.86$
& $88.09$\\
\ts{\algoname}(ours)
& $98.72$
& $97.56$
& $\bf{98.14}$
& $97.68$
& $97.39$
& $\bf{97.54}$
& $97.63$
& $98.30$
& $\bf{97.96}$\\
\bottomrule
\end{tabular}
\label{table:different_partition}
\end{table*}

\begin{table*}[h]
\centering
\caption{Normalized Decoding Accuracy results when coding over experts trained on different datasets. \ts{\algoname} shows a significant improvement in performance compared to baselines while being compute-efficient.}
\color{black}

\begin{tabular}{lccc|ccc|ccc}
\toprule
&
\multicolumn{3}{c}{MNIST + FashionMNIST} &
\multicolumn{3}{c}{CIFAR10 + FashionMNIST } &
\multicolumn{3}{c}{CIFAR10 + MNIST}\\
\cmidrule{2-10}
Algorithm & 
\multicolumn{1}{c}{MNIST} &
\multicolumn{1}{c}{FashionMNIST} &
\multicolumn{1}{c}{Avg.} &
\multicolumn{1}{c}{CIFAR10} &
\multicolumn{1}{c}{FashionMNIST} &
\multicolumn{1}{c}{Avg.} &
\multicolumn{1}{c}{CIFAR10} &
\multicolumn{1}{c}{MNIST} &
\multicolumn{1}{c}{Avg.}\\
\midrule
Vanilla Averaging \cite{wortsman2022model}
& $42.33$
& $73.27$
& $57.80$
& $66.65$
& $80.12$
& $73.39$
& $86.08$
& $68.84$
& $77.46$\\
Task Arithmetic \cite{IlharcoRWSHF23}
& $52.96$
& $80.99$
& $66.98$
& $76.83$
& $84.25$
& $80.54$
& $86.08$
& $68.84$
& $77.46$\\
RegMean \cite{wortsman2022model}
& $73.18$
& $78.83$
& $76.00$
& $87.01$
& $86.62$
& $86.81$
& $83.20$
& $69.40$
& $76.30$\\
Ensemble Distillation
& $82.63$
& $75.27$
& $78.95$
& $62.44$
& $70.38$
& $66.41$
& $65.56$
& $87.29$
& $76.42$\\
\ts{\algoname}(ours)
& $80.36$
& $83.99$
& $\bf{82.17}$
& $89.12$
& $85.86$
& $\bf{87.49}$
& $92.89$
& $84.34$
& $\bf{88.62}$\\
\bottomrule
\end{tabular}
\label{table:different_datasets}
\end{table*}

\section{Experiments} \label{sec:experiments}
In this section we demonstrate the effectiveness of \ts{\algoname} for erasure coding on neural networks trained on real-world vision datasets. To do so, we first introduce the metric that we use in our experiments.

\textbf{Normalized Decoding Accuracy.} Recall in \Cref{sec:problemFormulation} we use $f_{\btheta_i}(\bx)$ to denote the outputs of the $i$-th neural network and $\rcoi(\bx)$ to denote the decoded outputs for network $i$ for some given coded model $\btheta$ (see \Cref{eq:decoded_output}). Let $\mathcal{S}_i^{\text{test}}$ be the test data associated with neural network $i$. We define the \textit{Normalized Decoding Accuracy} (NDA) for the $i$-th network as follows:
\begin{align}
    & 100 \times \frac{\sum_{(\bx,\by) \in \mathcal{S}_i^{\text{test}}} \ind{\argmax\left(\rcoi(\bx)\right) = y}}{\sum_{(\bx,\by) \in \mathcal{S}_i^{\text{test}}} \ind{\argmax \left(f_{\btheta_i}(\bx)\right) = y}}
    \label{eq:nda_def}
\end{align}
where $\ind{\cdot}$ is the indicator function. We see that the numerator of \Cref{eq:nda_def} measures the accuracy of the decoded outputs while the denominator measures the accuracy of the original network $f_{\btheta_i}(\cdot)$. For linear models since $\rcoi(\bx) = f_{\btheta_i}(\bx)$, this ratio would always be $100$; however for non-linear networks, due to the approximations introduced in the coding step, i.e., $\rcoi(\bx) \approx f_{\btheta_i}(\bx)$, this ratio is usually less than $100$. Thus, $(100- \text{NDA})$ gives us a measure of the error introduced by the approximate erasure coding over non-linear models.

\textbf{Baselines.} We compare $\ts{\algoname}$ with $4$ other baselines including $3$ which are adopted from the model merging literature. Vanilla Averaging \cite{wortsman2022model} is the first and most common baseline in model merging literature where the merged/coded model is constructed by a simple weighted average of the parameter vectors of the individual
models, i.e., $\btheta = \sum_{i=1}^n \beta_i \btheta_i$. Next, we compared with Task Arithmetic \cite{IlharcoRWSHF23}, where the coded model is constructed as $\btheta = \btheta_0 + \alpha\sum_{i=1}^n (\btheta_i - \btheta_0)$ with $\btheta_0$ being our base foundation model and $\alpha$ being a hyperparameter which is tuned using validation data. RegMean \cite{Jin0P023} is a recently proposed state-of-the-art model fusion method which uses the Gram matrices of the data for model fusion. Lastly, we also compare with the Ensemble Distillation baseline, as outlined in \Cref{sec:problemFormulation}. 

\textbf{Experimental Setup.}
We use a ResNet50 pretrained on ImageNet as our foundation model. The datasets we consider are MNIST, FashionMNIST and CIFAR10, all of which consist of $10$ classes, i.e., $K = 10$. In all experiments, we set the number of coded models $n = 2$ and coding coefficients to be $\beta_1 = 0.5$ and $\beta_2 = 0.5$ for simplicity. Now to simulate $n = 2$ experts, each of which specializes in a particular type of query, we consider the following two settings. In the first case we consider experts that are trained on different partitions of the same dataset. We split the given dataset into two partitions $\mathcal{S}_1$ and $\mathcal{S}_2$ where $\mathcal{S}_1$ consists of all the data corresponding to labels $\{1,2,\dots,5\}$ and $\mathcal{S}_2$ consists of the data corresponding to labels $\{6,7\dots, 10\}$ and fine-tune a neural network on each partition. In the second case, we consider experts that are fine-tuned on different datasets itself, for e.g., where $\mathcal{S}_1$ is the CIFAR-10 dataset and $\mathcal{S}_2$ is the MNIST dataset. 
For algorithms which require access to data to create the coded model (RegMean, Ensemble Distillation, \ts{\algoname}), we sample $P' = 100$ datapoints from both $\mathcal{S}_1$ and $\mathcal{S}_2$ giving us $P = 200$ datapoints in total, which is less than $1\%$ of the total data in $\mathcal{S}_1$ and $\mathcal{S}_2$. Additional details and an ablation study evaluating the effect of $P$ on the normalized decoding accuracy can be found in the Appendix.

\textbf{Discussion.} \Cref{table:different_partition} shows the normalized decoding accuracy results when coding over experts trained on different partitions of the same dataset while \Cref{table:different_datasets} shows the results of coding over experts trained on different datasets for different combinations of datasets.
In all cases we see that \ts{\algoname} achieves the highest average normalized decoding accuracy while avoiding any expensive computational procedures such as distillation (Ensemble Distillation) or computing the Gram matrix of data (RegMean). Specifically for \Cref{table:different_partition} we see that \ts{\algoname} is the only algorithm which consistently achieves greater than $97.5\%$ average normalized decoding accuracy which implies that there is a less than $2.5\%$ loss in accuracy compared to the individual models. In \Cref{table:different_datasets} we see that there is a larger drop in accuracy when coding over models trained on different datasets which can be attributed to the greater data heterogeneity used to fine-tune the respective models. Nonetheless, \ts{\algoname} continues to outperform baselines with almost $10\%$ in some cases like CIFAR10+MNIST.

\section{Concluding Remarks} \label{sec:conclusion}
In this paper, we propose \ts{\algoname}, an algorithm that leverages erasure coding for multi-model neural network inference using an equivalence with the KL barycenter problem in its design. 
% Our method can flexibly handle heterogeneous and time-varying inference traffic on neural network models that are deployed on shared cloud-based shared infrastructure. 
Our solution is both efficient in resource utilization (needs less than $1\%$ of training data) and avoids any expensive computational procedures such as ensemble distillation. We demonstrate via experiments over that our method significantly improves decoding accuracy compared to baselines when coding over neural networks trained on real-world vision datasets in various settings. Directions for future work include characterizing the performance on a wider range of model architectures such as transformers and coding over a larger set of models.

\section*{Acknowledgment}
This work was supported in part by NSF CCF 2045694, CNS-2112471, CPS-2111751, and ONR N00014-23-1-2149, and the CMU Benjamin Garver Lamme/Westinghouse Fellowship. The authors would also like to thank Tuhinangshu Choudhury for insightful discussions.

\bibliographystyle{IEEEtran}
\bibliography{references}

\newpage

\onecolumn
\begin{center}
    {\LARGE\textbf{{Appendix}}} 
    \vspace{0.5cm}
\end{center}

\section{Additional Experimental Details}

We use PyTorch to run all our experiments. 
For fine-tuning we use the AdamW optimizer with a learning rate of $10^{-5}$, batch size of $128$ and weight decay $0.1$. For MNIST dataset we fine-tune for $1$ epoch, for FashionMNIST we fine-tune for $5$ epochs and for CIFAR10 we fine-tune for $3$ epochs. During fine-tuning, we freeze the BatchNorm parameters of the model. We find that while this does not affect the fine-tuning accuracy it significantly improves the normalized decoding accuracy for all algorithms. A more extensive evaluation on the effect of using BatchNorm for erasure coding is left as future work. For Task Arithmetic we use the available $P$ samples as the validation data and tune $\alpha$ in the range $[0.05, 0.1, 0.15,\dots, 1.0]$ to find the $\alpha$ which achieves the highest normalized decoding accuracy on the validation data. For \ts{\algoname}, we similarly tune $\lambda$ in the range $[10^{-5}, 10^{-4}, \dots, 1]$ using the $P$ samples as validation data. To implement RegMean we use the code publicly available on the official repository on Github. For Ensemble Distillation we again use the AdamW optimizer with a learning rate of $10^{-5}$, batch size of $8$, weight decay $0.1$ and run the optimization for $20$ epochs.

\section{Additional Experiments and Results}

\begin{figure}[h]
  \centering
   \subfloat[]{\includegraphics[width=0.4\linewidth]{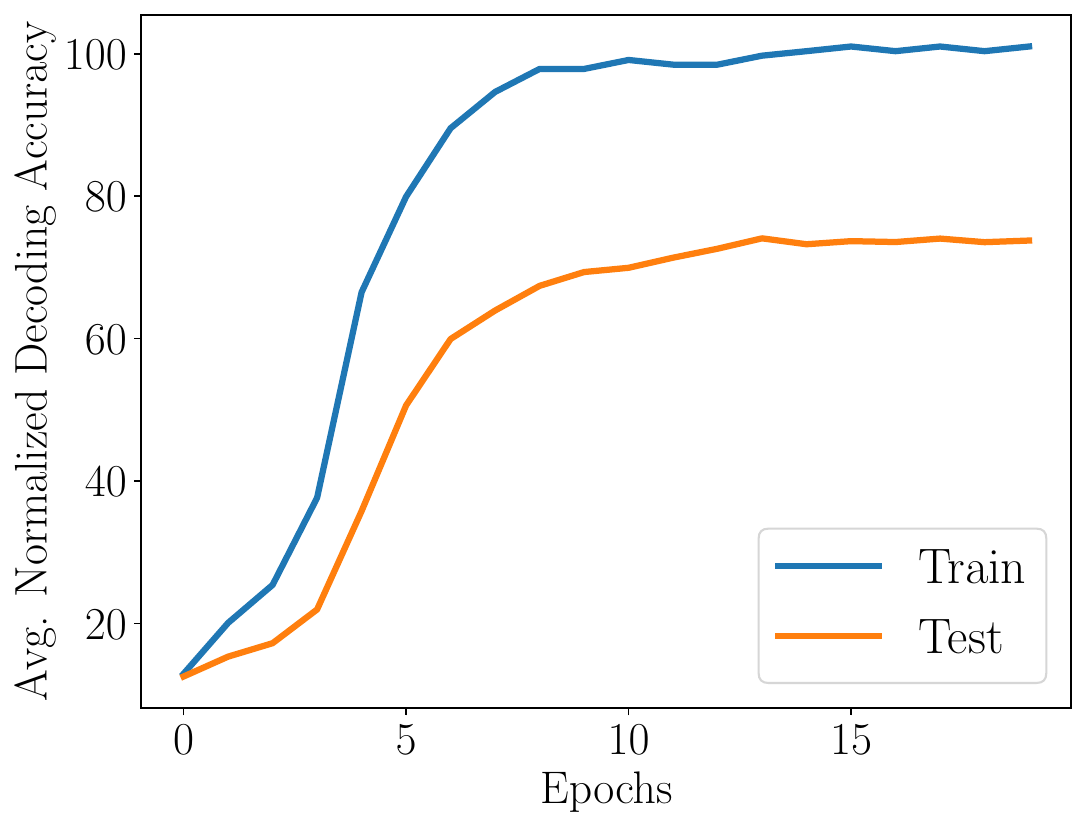}}\label{fig:acc_vs_epochs}
   \hspace{5em}
   \subfloat[]{\includegraphics[width=0.4\columnwidth]{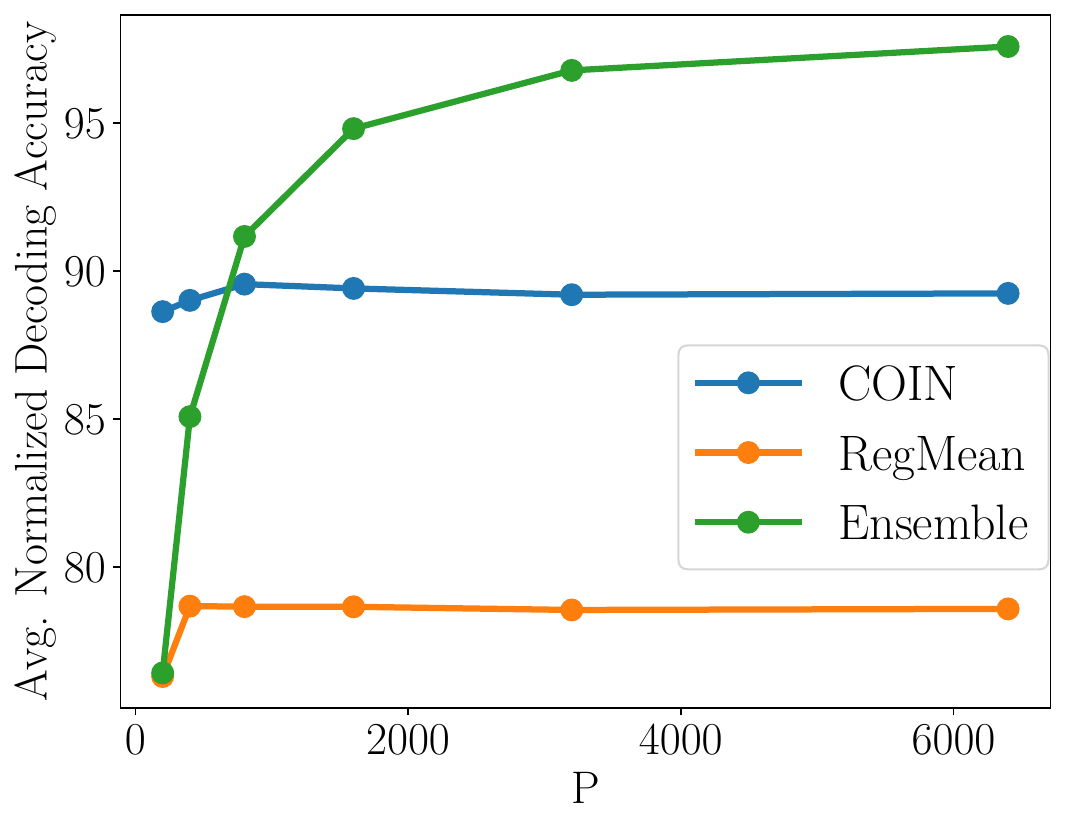}}
   \caption{ (a) shows the average normalized decoding accuracies computed on the train set and test set for the Ensemble Distillation baseline as a function of the number of optimization epochs when coding over networks trained on CIFAR-10 and MNIST. The accuracy on the train set reaches close to $100$ but accuracy on test set saturates close to $75$, implying overfitting. (b) shows the average normalized decoding accuracy for \ts{\algoname}, RegMean and Ensemble Distillation in the same setting as a function of the number of datapoints $P$. We see only a slight increase in the accuracy of \ts{\algoname} as we increase $P$, which demonstrates the data-efficiency of our approach. }
  \label{fig:overfitting_and_effect_of_P}
\end{figure}

We conduct additional experiments in the setting where we are coding over neural networks fine-tuned on CIFAR-10 and MNIST respectively to showcase the overfitting behavior of the Ensemble Distillation baseline and the effect of the number of datapoints $P$ on the decoding accuracy. \Cref{fig:overfitting_and_effect_of_P}(a) shows the average normalized decoding accuracies computed on the train set and test set for the Ensemble Distillation baseline as we train the coded model. We see that while the decoding accuracy for the train set quickly reaches close to $100$, the accuracy for the test set saturates close to $75$, implying that the coded model is clearly overfitting the training set. Note that we are using a weight decay of $0.1$ in the optimization procedure which is a standard technique to prevent overfitting. \Cref{fig:overfitting_and_effect_of_P}(b) shows the average normalized decoding accuracy for \ts{\algoname}, RegMean and Ensemble Distillation as a function of the number of datapoints $P$. We see that as $P$ increases, the performance of Ensemble Distillation improves significantly, which is expected since the coded model is less likely to overfit as the size of the training data increases. Nonetheless, we note that the cost of computing the coded model using Ensemble Distillation also grows significantly as $P$ increases. On the other hand there is only a slight improvement in the accuracy for \ts{\algoname} which reinforces the data efficiency and implicit computational ease of our proposed method. RegMean also sees an improvement as we increase the number of samples $P$ which can be attributed to a more accurate estimation of the Gram matrices of the data.

\end{document}